\documentclass[10pt,twocolumn,letterpaper]{article}

\usepackage{cvpr}

\usepackage[T1]{fontenc}

\usepackage{cuted}

\usepackage{currfile}

\usepackage{caption}

\usepackage{multirow}
\usepackage{tikz}
\usepackage{algorithm}
\usepackage[noend]{algpseudocode}
\usetikzlibrary{arrows.meta,calc,decorations.pathreplacing}
\usepackage{pgfplots}
\pgfplotsset{compat=1.17}
\usepgfplotslibrary{fillbetween}
\newcommand{\methodstyled}{\textcolor{cvprblue}{\textbf{\texttt{MemJev}}}\xspace}
\newcommand{\method}{MemJev\xspace}
\newcommand{\jev}{Jev\xspace}
\newcommand{\groundsg}{GroundSG\xspace}

\definecolor{cvprblue}{rgb}{0.21,0.49,0.74}
\usepackage[pagebackref,breaklinks,colorlinks,allcolors=cvprblue]{hyperref}

\def\confName{CVPR}
\def\confYear{2027}

\title{\methodstyled: Training Jev for Hierarchical Action Policy\\ with Recurrent Visual Memory}

\author{
  Jangho Park$^{1}$ \qquad
  Beomsu Kim$^{1}$ \qquad
  Gihyun Kwon$^{2,*,\dagger}$ \qquad
  Jong Chul Ye$^{1,\dagger}$ \\[4pt]
  $^{1}$KAIST AI \qquad $^{2}$Amazon \\
  {\small\bfseries \url{https://memjev.github.io/}} \\
}

\begin{document}
\maketitle
{\renewcommand{\thefootnote}{$*$}\footnotetext{Work done outside of Amazon.}}
{\renewcommand{\thefootnote}{$\dagger$}\footnotetext{Co-corresponding authors.}}
\begin{abstract}
History-dependent embodied tasks require robots to remember past observations when choosing the next subgoal and target object. Once such a decision is made, a pretrained action policy can execute it from the current observation without retraining. The remaining challenge is to maintain and efficiently query a compact representation of the growing visual history at bounded cost. We present \methodstyled, a high-level memory system built on an open-source vision-language variant of \jev, a decision-making model. We train it to keep the visual history in a fixed-size recurrent state, and it scores candidate subgoals and objects in a single forward pass. Decision questions read temporary copies of this state alongside recent frames, keeping persistent history separate from decision queries. With fixed recent-window and candidate budgets, history storage and history-dependent computation per decision remain bounded independently of episode length. We train the model on decision questions and memory questions derived from RoboMME demonstrations. Across all 16 RoboMME tasks with a frozen low-level policy, \method achieves 57.3\% mean success over three seeds, compared with 33.8\% for Qwen-VL subgoals and 42.5\% for MemER under the same policy and test episodes, while taking a median of 158\,ms per decision, against more than 1\,s for both methods, and keeping the history in a fixed-size memory as the episode grows longer.
\end{abstract}
\begin{figure}[t]
  \centering
  \captionsetup{skip=3pt}
  \resizebox{\columnwidth}{!}{%
\begin{tikzpicture}[x=1cm,y=1cm, font=\sffamily\scriptsize,
  box/.style={draw=black!45,rounded corners=2pt,fill=black!3,
    minimum height=.57cm,align=center,inner sep=3pt},
  flow/.style={-{Stealth[length=1.6mm]},thick,black!65},
  pics/flame/.style={code={
    \path[fill=orange!90!red,draw=orange!70!red,line width=.25pt]
      (0,-.13) .. controls (-.22,-.09) and (-.15,.08) .. (-.08,.14)
      .. controls (-.09,.03) and (.03,.04) .. (0,.28)
      .. controls (.21,.11) and (.22,-.09) .. (0,-.13);
    \path[fill=yellow!80] (0,-.09) .. controls (-.09,-.04) and (-.04,.03) .. (.02,.08)
      .. controls (.01,.01) and (.12,-.04) .. (0,-.09);
  }},
  pics/historyfilm/.style={code={
    \path[draw=black!65,fill=black!60,rounded corners=.7pt]
      (-.53,-.23) rectangle (.53,.23);
    \foreach \x in {-.47,-.31,-.15,.01,.17,.33,.49} {
      \fill[white] (\x-.025,.165) rectangle (\x+.025,.205);
      \fill[white] (\x-.025,-.205) rectangle (\x+.025,-.165);
    }
    \foreach \x in {-.34,0,.34} {
      \fill[white] (\x-.145,-.135) rectangle (\x+.145,.135);
      \fill[black!18] (\x-.13,-.12) -- (\x-.04,.025)
        -- (\x+.025,-.045) -- (\x+.08,.055) -- (\x+.13,-.12) -- cycle;
      \fill[black!35] (\x-.075,.075) circle (.025);
    }
    \node[font=\sffamily\scriptsize,anchor=south,inner sep=0pt] at (0,.29) {History};
  }},
  pics/snow/.style={code={
    \foreach \a in {0,60,...,300} {
      \draw[cyan!70!blue,line width=.65pt,rotate=\a] (0,0)--(0,.2)
        (-.055,.12)--(0,.075)--(.055,.12);
    }
  }}]
\path[use as bounding box] (-.15,-.8) rectangle (8.95,6.6);
\fill[black!3,rounded corners=5pt] (-.08,4.5) rectangle (8.88,6.52);
\fill[black!3,rounded corners=5pt] (-.08,2.23) rectangle (8.88,4.3);
\fill[cvprblue!6,rounded corners=5pt] (-.08,-.17) rectangle (8.88,2.05);
\node[anchor=west,font=\sffamily\bfseries\small] at (.12,6.2)
  {(a) Memory-augmented action policy};
\node[minimum width=1.06cm,minimum height=.46cm,inner sep=0pt] (h1) at (.85,5.38) {};
\pic at (h1) {historyfilm};
\node[box,fill=white,minimum width=1.6cm] (m1) at (4.25,5.38) {Policy memory};
\node[box,fill=white,minimum width=1.45cm] (p1) at (7.65,5.38) {Action policy};
\draw[flow] (h1)--(m1); \draw[flow] (m1)--(p1);
\pic at ([xshift=-1pt,yshift=2pt]m1.north east) {flame};
\pic at ([xshift=-1pt,yshift=2pt]p1.north east) {flame};
\node[anchor=west,text=black!65] at (.12,4.78) {Memory-conditioned actions};

\node[anchor=west,font=\sffamily\bfseries\small] at (.12,3.98)
  {(b) Keyframe-retention planner};
\node[minimum width=1.06cm,minimum height=.46cm,inner sep=0pt] (h2) at (.85,3.13) {};
\pic at (h2) {historyfilm};
\path[draw=black!25,fill=white,rounded corners=2pt] (2.29,2.62) rectangle (4.03,3.32);
\path[draw=black!35,fill=white,rounded corners=2pt] (2.21,2.70) rectangle (3.95,3.40);
\node[box,fill=white,text width=1.43cm,inner sep=0pt,minimum width=1.74cm,minimum height=.70cm] (m2) at (3.0,3.13) {Selected\\keyframes};
\node[box,fill=white,minimum width=1.45cm] (p2) at (5.35,3.13) {\textbf{VLM} subgoal\\generation};
\node[box,fill=white,minimum width=1.45cm] (c2) at (7.65,3.13) {Action policy};
\draw[flow] (h2)--(m2); \draw[flow] (m2)--(p2); \draw[flow] (p2)--(c2);
\pic at ([xshift=-1pt,yshift=2pt]p2.north east) {flame};
\pic at ([xshift=-1pt,yshift=2pt]c2.north east) {flame};
\node[anchor=west,text=black!65] at (.12,2.43) {Retain frames; generate subgoals};

\begin{scope}[yshift=-.15cm]
\node[anchor=west,font=\sffamily\bfseries\small,text=cvprblue] at (.12,1.99)
  {(c) MemJev (ours)};
\path[draw=cvprblue!45,fill=white,rounded corners=4pt]
  (2.03,.47) rectangle (6.33,1.68);
\node[box,draw=cvprblue,fill=cvprblue!9,minimum width=1.43cm] (m3) at (3.0,.88) {Recurrent\\state};
\node[box,draw=cvprblue,fill=cvprblue!9,minimum width=1.45cm] (p3) at (5.35,.88) {\textbf{Jev} subgoal\\scoring};
\node[minimum width=1.06cm,minimum height=.46cm,inner sep=0pt] (h3) at (.85,.88) {};
\pic at (h3) {historyfilm};
\node[box,fill=white,minimum width=1.45cm] (c3) at (7.65,.88) {Action policy};
\draw[flow] (h3)--(m3); \draw[flow,draw=cvprblue] (m3)--(p3); \draw[flow] (p3)--(c3);
\draw[-{Stealth[length=1.2mm]},cvprblue,thick]
  ([xshift=-5pt]m3.north) .. controls (2.65,1.62) and (3.35,1.62) .. ([xshift=5pt]m3.north);
\pic at ([xshift=-1pt,yshift=2pt]m3.north east) {flame};
\pic at ([xshift=-1pt,yshift=2pt]p3.north east) {flame};
\pic at ([xshift=-1pt,yshift=2pt]c3.north east) {snow};
\node[anchor=west,text=cvprblue] at (.12,.22) {Compress history; score subgoals};
\node[font=\sffamily\scriptsize,text=cvprblue,fill=cvprblue!6,inner sep=1pt]
  at (5.0,1.99) {High-level memory model};
\end{scope}
\pic at (.2,-.55) {flame};
\node[anchor=west] at (.45,-.52) {Trainable module};
\pic at (3.25,-.52) {snow};
\node[anchor=west] at (3.55,-.52) {Frozen module};
\end{tikzpicture}%
}
  \caption{\textbf{Memory interfaces for action policies.}
  Policy-level memory conditions action generation; keyframe planners retain observations for subgoal prediction. \method compresses history recurrently and scores candidates above a frozen policy.}
  \label{fig:memory_comparison}
  \vspace{-10pt}
\end{figure}
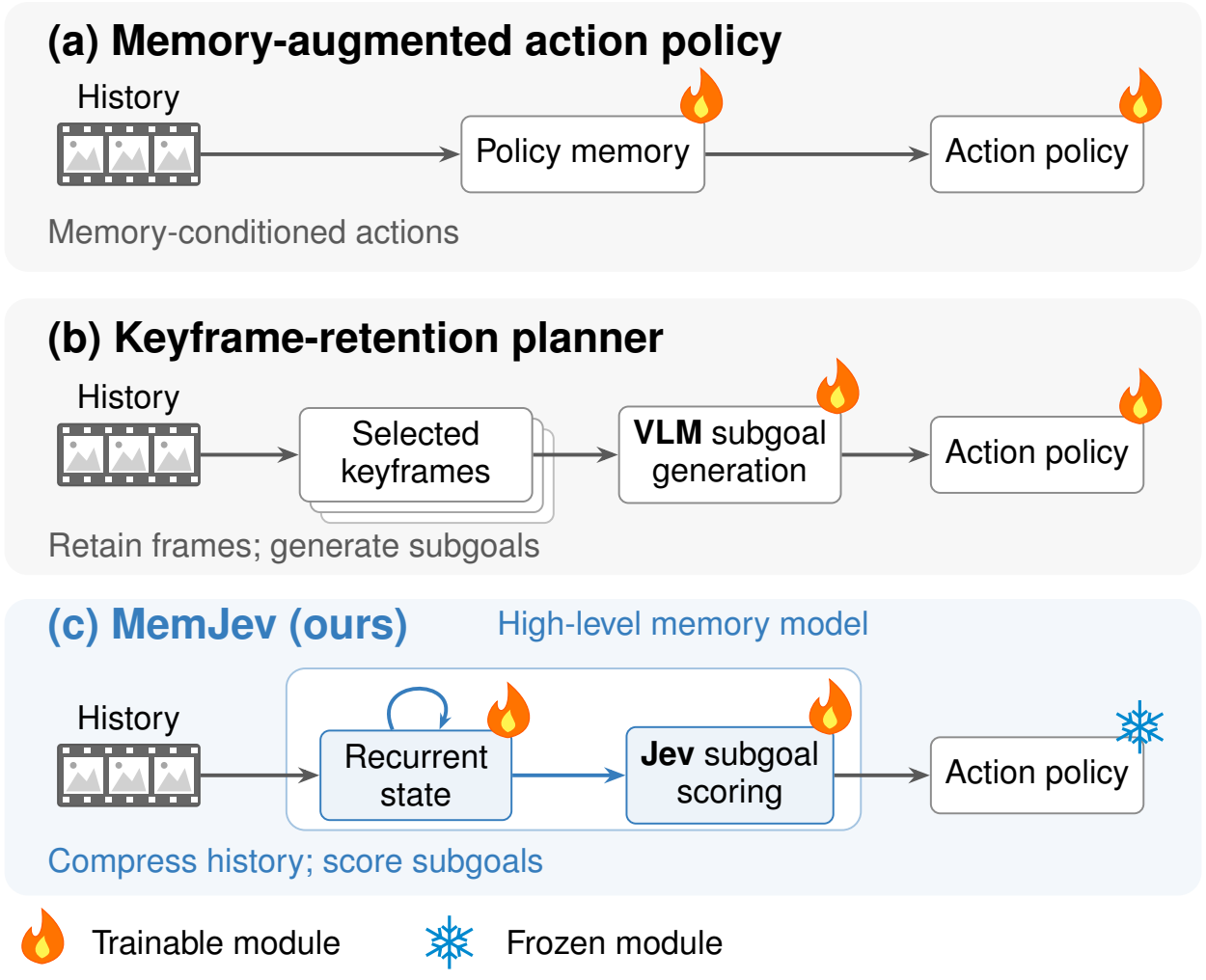

\section{Introduction}
\label{sec:intro}

Vision-language-action (VLA) models have broadened the range of embodied tasks that robots can perform from language instructions~\cite{brohan2023rt2,kim2024openvla,black2024pi0,intelligence2025pi05}.
Many of these policies nevertheless rely on the current observation or a short visual history, limiting their use when the information needed for a decision is no longer visible.
For example, a cube is initially hidden beneath one of several identical cups, which are then shuffled while the cube remains concealed.
To retrieve the cube, the robot must remember which cup originally covered it and track that cup through the shuffle.
Likewise, deciding whether to repeat an action or proceed to the next step can depend on how many repetitions have already been completed.
The current observation alone may therefore be insufficient to determine the correct action: the robot must retain relevant information from earlier events~\cite{dai2026robomme}.

Memory-augmented policies condition action generation on historical information through recurrent states, retrieved memory representations, or multimodal summaries.
Examples include recurrent fast-weight memory in RoboTTT~\cite{jiang2026robottt} and T$^2$Mem~\cite{liu2026t2mem}, perceptual-cognitive memory banks in MemoryVLA~\cite{shi2026memoryvla}, and complementary short-term visual and long-term language memory in MEM~\cite{torne2026mem}.
These methods train memory jointly with action generation, requiring substantial computation and time to update a large action policy.
However, when history affects high-level decisions rather than action generation, the action policy itself need not be updated.
In the shuffled-cup example, history identifies which cup should be lifted, and this high-level decision can be passed to a pretrained action policy as an instruction and its current image location.
This motivates a hierarchical interface that separates history-dependent subgoal selection from low-level execution~\cite{ahn2022saycan,shi2025hirobot,dai2026robomme}, allowing supervision to focus on memory and high-level decisions while keeping the policy fixed.

When the low-level policy acts without access to history, the planner must retain the information needed to select successive grounded subgoals. MemER~\cite{sridhar2025memer} learns to nominate and retain keyframes for subsequent subtask prediction, while PrediMem~\cite{lei2026robomemarena} maintains recent and keyframe buffers and uses a predictive-coding objective. These approaches use a high-level VLM to predict the current subtask from visual memory. In MemER, the high-level model runs at approximately 1\,Hz, compared with approximately 2\,Hz for the action policy~\cite{sridhar2025memer}. On RoboMME, external-VLM inference also introduces substantial overhead: generating subgoals with Qwen-VL requires roughly three times the computation of $\pi_{0.5}$, while MemER requires nearly five times~\cite{dai2026robomme}.
Keyframe-based memory also introduces a representational limitation: it encodes task history as sparse snapshots. Reconstructing progress from these snapshots can be difficult when repeated steps appear visually similar. A recurrent state offers an alternative by incrementally summarizing observations. Separately, direct candidate scoring avoids generating each subgoal token by token.
Taken together, these observations raise two questions. \emph{Can we keep the action policy frozen and improve success on history-dependent tasks by training a high-level model with fixed-size recurrent memory?} \emph{Can a vision-language model drive the policy by selecting among candidate decisions instead of generating each subgoal as full text?}

\begin{figure*}[t]
  \centering
  \input{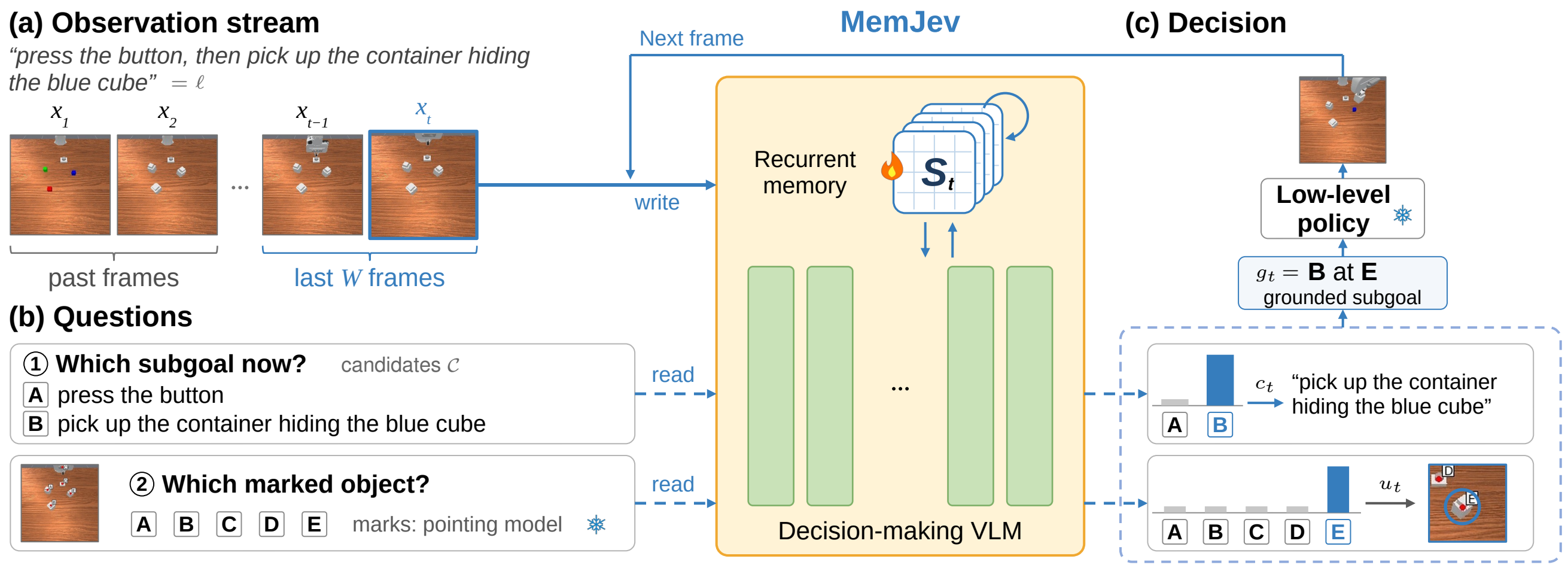}
  \caption{\textbf{Overview of \method.}
  (a) Every frame updates the recurrent state $S_t$, and only the last $W$ frames stay visible to attention; (b) a subgoal question and, when a changed subgoal requires a target, an object question read temporary state copies; (c) the selected subgoal $c_t$ and object location $u_t$ form the grounded subgoal $g_t$ for the frozen low-level policy.}
  \label{fig:pipeline}
  \vspace{-10pt}
\end{figure*}

We introduce \method, a high-level memory system built on an open-source decision-making VLM~\cite{vev2026}, a variant of \jev~\cite{typesafe2026jev}.
\method answers the first question with recurrent visual memory and a training pipeline for it, and the second with non-autoregressive candidate scoring (\cref{fig:pipeline}).
The recurrent visual memory compresses each incoming frame into a fixed-size recurrent state, so neither the memory nor the computation per decision grows with the episode, while the action policy stays frozen.
We also propose a training pipeline that enables a decision-making model to operate on memory-dependent embodied tasks by converting robot demonstrations into decision questions and memory questions without manual labeling.
Because attention covers only the most recent frames, questions about earlier events can be answered only through the recurrent state.
At each decision, \method scores all candidate subgoals and target objects in one forward pass per question instead of generating the subgoal as text, making most decisions several times faster than text generation.
Our contributions are as follows.
\begin{itemize}
  \setlength\itemsep{1pt}
  \item We propose \method, in which a decision-making VLM with recurrent visual memory selects grounded subgoals for a frozen low-level policy.
  \item We propose a training method that gives a decision-making model memory, with decision and memory questions built from robot demonstrations.
  \item \method reaches 57.3\% mean success on RoboMME, 14.8 points above the best method with the same low-level policy, and its per-decision memory cost stays constant while theirs grows with the episode.
\end{itemize}

\section{Related Work}
\label{sec:related}

\paragraph{Memory-augmented action policies.}
Historical information can condition action generation through explicit visual context~\cite{jang2025contextvla,koo2026hamlet}, memory banks, semantic summaries, recurrent states, or previously predicted actions~\cite{zhao2023learning,liu2025bidirectional,black2025real,kwon2026don}.
SAM2Act+~\cite{fang2025sam2act} and MemoryVLA~\cite{shi2026memoryvla} use memory banks for spatial and perceptual-cognitive information, respectively, while MEM~\cite{torne2026mem} combines short-term visual history with long-term language memory.
Embodied-SlotSSM~\cite{chung2026embodiedslotssm} maintains object-centric histories through slot-state-space modeling, while RoboTTT~\cite{jiang2026robottt} and T$^2$Mem~\cite{liu2026t2mem} incorporate fast-weight memory into action prediction.
RoboMME~\cite{dai2026robomme} compares memory representations and integration strategies on a shared policy backbone.
When history determines the next subgoal and its target object, grounded subgoals can convey this information to a pretrained policy.
\method learns memory for these decisions while preserving action capability.

\paragraph{Memory-augmented hierarchical control.}
Hierarchical systems connect high-level decisions to low-level execution through language instructions~\cite{ahn2022saycan,shi2025hirobot} or intermediate language motions~\cite{belkhale2024rth}.
History-dependent planners must retain evidence for subsequent subtasks and targets.
Inner Monologue~\cite{huang2023innermonologue} accumulates scene descriptions and execution feedback in a language-model planning context, providing an explicit semantic history.
MemER~\cite{sridhar2025memer} and PrediMem~\cite{lei2026robomemarena} instead learn to retain historical keyframes and use them alongside recent observations for subtask prediction.
Keyframes preserve visual evidence for later reasoning, and the context of the planner grows with the evidence that is kept.
Our method compresses observations into a fixed-size recurrent state trained for grounded subgoal selection.

\paragraph{Recurrent memory.}
Linear attention admits a recurrent formulation~\cite{katharopoulos2020linear}, while selective state-space models such as Mamba~\cite{gu2024mamba} use input-dependent dynamics to propagate or forget information without retaining the token history.
Delta-rule models maintain a matrix-valued associative memory for each recurrent layer and head, which incoming keys and values update and queries read, so the state size does not depend on sequence length~\cite{yang2024deltanet,yang2025gdn}.
DeltaNet~\cite{yang2024deltanet} revises key--value associations with error-correcting updates, and Gated DeltaNet~\cite{yang2025gdn}, Kimi Delta Attention~\cite{kimi2025linear}, and Gated DeltaNet-2~\cite{hatamizadeh2026gdn2} add forgetting, channel-wise decay, and separate erase and write gates, respectively, whereas test-time training~\cite{sun2024ttt} and MesaNet~\cite{vonoswald2025mesanet} update the state by optimizing an objective on the incoming sequence.
We apply recurrent sequence modeling to visual history at the high-level decision interface.

\paragraph{Subgoal generation and selection.}
High-level decisions can be obtained through candidate selection or instruction generation.
SayCan~\cite{ahn2022saycan} selects predefined skills using language-model likelihoods and affordance scores, whereas MemER~\cite{sridhar2025memer} and PrediMem~\cite{lei2026robomemarena} generate subtask instructions from visual history.
\method{} utilizes a decision-making VLM to score candidate subgoals and objects conditioned on recurrent visual memory in a single forward pass per question, avoiding autoregressive instruction decoding.
\section{MemJev}
\label{sec:method}

\method separates the use of visual history from low-level action prediction.
A decision-making VLM with recurrent memory selects a subgoal at every decision step and passes it to a frozen low-level policy for execution (\cref{fig:pipeline}).
Two further components supply the options of its decisions: a subgoal predictor $\rho_\psi$ lists the candidate subgoals $\mathcal{C}$ of an instruction, and a pointing model proposes the object locations in the current frame.

\subsection{Preliminaries}
\label{sec:method:setting}

\paragraph{Problem formulation.}
We consider language-conditioned robot tasks in which the correct action depends on events that are no longer visible, such as which container was placed over a cube or how many times a motion has already been performed~\cite{dai2026robomme}.
We adopt a hierarchical formulation~\cite{shi2025hirobot,sridhar2025memer,dai2026robomme}.
The instruction $\ell$ is a text prompt that states the goal of the task.
A subgoal $c$ is a text that describes one of the steps required to accomplish $\ell$.
A grounded subgoal $g=(c,u)$ is a subgoal together with the image location $u=\langle y,x\rangle$ of the object it refers to; $u=\varnothing$ when the subgoal refers to no object.
Given $\ell$ and the frames $x_{1:t}$ observed up to step $t$, a high-level model selects a grounded subgoal $g_t$, and a frozen low-level policy (\cref{sec:exp:setup}) executes it from the current observation $o_t$,
\begin{equation}
  A_t \sim \pi_{\mathrm{low}}\!\left(\,\cdot \mid o_t, \ell, g_t\right),
  \label{eq:low}
\end{equation}
where $A_t$ is the action chunk executed until the next decision.
The policy receives both $\ell$ and $g_t$ as text: $\ell$ is the same throughout the episode, and $g_t$ specifies the step to execute now.
The low-level policy sees only the current observation, so information from earlier frames can reach it only through $g_t$.
The high-level model therefore has to select $g_t$ from the whole history $x_{1:t}$, which grows with the length of the episode.

\paragraph{Decision-making model.}
\jev~\cite{typesafe2026jev} is a language-model-based decision-making model: given a question and a set of options, it returns a probability for each option instead of generating text.
We use an open-source vision-language variant of \jev~\cite{vev2026}, which we refer to as the decision-making VLM.
Given a context $X$ of images and text, a question $q$ and a set of options $\mathcal{Y}$, it returns
\begin{equation}
  p_\theta\!\left(y \mid X, q, \mathcal{Y}\right), \qquad y\in\mathcal{Y},
  \label{eq:jev}
\end{equation}
in one forward pass, as a softmax over the logits of the option letters at the last token.
Used directly for our problem, the model would be given the instruction and the whole history, $X=(\ell, x_{1:t})$, at every step, so the cost of a decision grows with the length of the episode.

\paragraph{Fast-weight memory.}
A fast-weight memory~\cite{ba2016fastweights,schlag2021fwp} stores a sequence in matrices of fixed size instead of keeping its tokens.
Each input writes to these matrices with a learned update rule and later inputs read from them.
We use such a memory for the frames of an episode and write $S_t$ for its content after frame $x_t$, which we call the state:
\begin{equation}
  S_t = f_\theta(S_{t-1}, x_t),
  \label{eq:state}
\end{equation}
where $S_0$ is the state after the instruction.
The update rule $f_\theta$ is Gated DeltaNet~\cite{yang2025gdn} (Appendix~\ref{app:gdn}), and the size of $S_t$ does not depend on $t$.
\method uses this state in place of the growing history in \cref{eq:jev}: \cref{sec:method:memory} trains the decision-making VLM to read frames older than a short window from $S_t$, and \cref{sec:method:inference} uses the trained model online to select $g_t$.

\subsection{Recurrent visual memory}
\label{sec:method:memory}

When every frame is visible to attention, a decision can be made from the frames directly, so nothing requires the state to hold earlier frames and the cost of a decision grows with the length of the episode.
In \method, the fast-weight state serves as a recurrent visual memory: frames older than a short window can influence a decision only through $S_t$, and the model is trained to answer under this constraint.

\paragraph{Temporally local attention.}
We make the attention of the decision-making VLM temporally local: a decision attends only to the instruction and the most recent $W$ frames, and all earlier frames are masked.
The fast-weight update, in contrast, receives every frame of the episode sequentially, so each frame is written to the state regardless of the window.
The update $S_t=f_\theta(S_{t-1},x_t)$ in \cref{eq:state} processes the $m$ tokens of frame $x_t$ through a recurrent update rule. Specifically, having initialized $\mathbf{M}_0=S_{t-1}$, the recurrent memory matrix is updated via
\begin{equation}
  \begin{aligned}
  \mathbf{M}_i &{}= \alpha_i \mathbf{M}_{i-1} + \beta_i\, \mathbf{k}_i\bigl(\mathbf{v}_i - \alpha_i \mathbf{M}_{i-1}^{\top} \mathbf{k}_i\bigr)^{\!\top},
  \end{aligned}
  \label{eq:write}
\end{equation}
for $i=1,\dots,m$, after which $S_t=\mathbf{M}_m$. This recurrence applies to the corresponding matrix in each recurrent layer and head, whose indices are omitted. The key $\mathbf{k}_i$, value $\mathbf{v}_i$ and gates $\alpha_i,\beta_i\in(0,1)$ are computed from token $i$ of frame $x_t$ (Appendix~\ref{app:gdn}).
A frame older than $W$ frames therefore reaches a decision only through the state, and \cref{eq:jev} becomes
\begin{equation}
  p_\theta\!\left(y \mid S_t, x_{t-W+1:t}, \ell, q, \mathcal{Y}\right), \qquad y\in\mathcal{Y},
  \label{eq:answer}
\end{equation}
whose input has a fixed size.

\begin{algorithm}[t!]
  \caption{Training of \method}
  \label{alg:training}
  \footnotesize
  \begin{algorithmic}[1]
    \Require training questions $(\ell, x_{1:t}, q, \mathcal{Y}, y^{\star})$ (\cref{sec:data}); window $W$
    \State train $\rho_\psi$ separately: minimize $-\log \rho_\psi(\mathcal{C}\mid \ell)$ \Comment{subgoal predictor}
    \For{each training step} \Comment{decision-making VLM}
      \State sample $(\ell, x_{1:t}, q, \mathcal{Y}, y^{\star})$
      \State $S_0 \gets$ state after reading $\ell$
      \For{$\tau=1,\dots,t$} \Comment{all $t$ frames update the state}
        \State $S_\tau \gets f_\theta(S_{\tau-1}, x_\tau)$, with attention on $\ell$ and $x_{\tau-W+1:\tau}$
      \EndFor
      \State $\mathcal{L} \gets -\log p_\theta(y^{\star} \mid S_t, x_{t-W+1:t}, \ell, q, \mathcal{Y})$ \Comment{\cref{eq:loss}}
      \State update the LoRA parameters of $\theta$ with $\nabla_\theta \mathcal{L}$, backpropagated through $S_1,\dots,S_t$
    \EndFor
  \end{algorithmic}
\end{algorithm}

\paragraph{Training objective.}
\label{sec:method:training}
A training example $(\ell, x_{1:t}, q, \mathcal{Y}, y^{\star})$ consists of the instruction $\ell$, the frames $x_{1:t}$ of a training episode up to frame $t$, a question $q$ asked at that frame, its options $\mathcal{Y}$, and the correct option $y^{\star}\in\mathcal{Y}$ (\cref{sec:data}).
We minimize the cross-entropy of the correct option,
\begin{equation}
  \mathcal{L}(\theta) = -\log p_\theta\!\left(y^{\star} \mid S_t, x_{t-W+1:t}, \ell, q, \mathcal{Y}\right).
  \label{eq:loss}
\end{equation}
Here $S_t$ is the state after all frames $x_{1:t}$, and $x_{t-W+1:t}$ are the last $W$ frames, which are the only frames visible to attention.
As summarized in \cref{alg:training}, each example is processed in one forward pass over $x_{1:t}$ and the question: the fast-weight update runs over all frames from the first one, and the attention layers apply the mask.
The gradient of $\mathcal{L}$ is backpropagated through all states $S_1,\dots,S_t$, so the update at every frame is trained and not only the one at frame $t$.
A question whose evidence lies more than $W$ frames back can only be answered through the state, so the model learns to store in the state what later questions will need.
We fine-tune the decision-making VLM with LoRA adapters~\cite{hu2021lora} on its language backbone and keep the vision encoder frozen (Appendix~\ref{app:training}).
The subgoal predictor $\rho_\psi$ of \cref{sec:method:inference} is trained separately on the instructions and subgoals of the training episodes, and the pointing model and the low-level policy are pretrained and kept frozen.

\subsection{Task execution with \method}
\label{sec:method:inference}
\label{sec:method:decisions}

At inference, \method processes an episode online: each incoming frame updates the state, and the subgoal at every decision step is obtained by evaluating \cref{eq:answer} on the current state and the most recent $W$ frames (\cref{alg:inference}).

\paragraph{Episode initialization.}
The decision-making VLM scores a given set of options, so at the start of an episode the subgoal predictor $\rho_\psi$ reads the instruction and returns the unordered set of subgoals that the episode may require,
\begin{equation}
  \mathcal{C} = \rho_\psi(\ell) = \{c^1,\dots,c^N\},
  \label{eq:cands}
\end{equation}
where a superscript indexes the elements of a set and a subscript indexes time.
When a task begins with a video demonstration, we sample one frame every $H$ control steps to obtain $x_{1:D}$. These frames update the state through \cref{eq:state} before the robot acts; $D=0$ for a task without a demonstration.

\paragraph{Episode execution.}
During execution, the model receives a front-camera frame $x_t$ every $H$ control steps. After each frame, the robot executes $H$ actions, one per control step, before the next frame $x_{t+1}$ arrives.
Every later frame $x_t$, $t>D$, is a decision step: the frame updates the state, our model selects the grounded subgoal $g_t$, and the low-level policy is called once and executes the $H$ control steps of its action chunk $A_t$, after which the next frame arrives.
The episode ends when the environment reports success or failure, or when the limit on control steps is reached (\cref{sec:exp:setup}).

\paragraph{Streaming inference.}
Between decisions \method keeps only $S_t$ and the attention cache of the instruction and the last $W$ frames.
If the tokens of a question were written to the state, every question would alter the memory that later decisions read.
A question is therefore a read-only query: its tokens are processed on a copy of the state and of the attention cache, and the copy is discarded after the answer.
The persistent state thus contains only the instruction and the frames, and the questions of one decision do not affect each other or later decisions.
Streaming inference is equivalent to the masked forward pass used in training, because the attention cache holds the tokens that the mask leaves visible.
The memory kept by our model therefore stays constant regardless of the length of the episode (\cref{tab:supp:cost}).
Decision latency is compared with other high-level models in \cref{sec:exp:latency}.

\begin{algorithm}[t!]
  \caption{Inference pipeline}
  \label{alg:inference}
  \footnotesize
  \begin{algorithmic}[1]
    \Require instruction $\ell$; demonstration video (optional)
    \State $\mathcal{C} \gets \rho_\psi(\ell)$ \Comment{subgoal candidates, once}
    \State $S_0 \gets$ state after reading $\ell$
    \If{a demonstration video is given}
      \State $x_{1:D} \gets$ frames of the video \Comment{one every $H$ control steps}
      \State $S_t \gets f_\theta(S_{t-1}, x_t)$ for $t=1,\dots,D$ \Comment{no action}
    \Else
      \State $D \gets 0$
    \EndIf
    \State $c_D \gets \varnothing$;\; $g_D \gets \varnothing$
    \For{$t=D+1, D+2, \dots$}
      \State receive $o_t$ with front frame $x_t$
      \State $S_t \gets f_\theta(S_{t-1}, x_t)$ \Comment{recurrent memory update}
      \State $c_t \gets$ \cref{eq:subgoal} \Comment{subgoal selection via VLM}
      \If{$c_t = c_{t-1}$}
        \State $g_t \gets g_{t-1}$ \Comment{location kept}
      \Else
        \State $g_t \gets (c_t, u_t)$, with $u_t$ from \cref{eq:object} or $\varnothing$
      \EndIf
      \State execute the $H$ control steps of $A_t \sim \pi_{\mathrm{low}}(\cdot \mid o_t, \ell, g_t)$
      \State stop if the episode has ended \Comment{success, failure or step limit}
    \EndFor
  \end{algorithmic}
\end{algorithm}

\begin{table*}[t]
  \centering
  \footnotesize
  \setlength{\tabcolsep}{2.0pt}
  \caption{\textbf{Success rates (\%) on the 16 RoboMME test tasks.} \textbf{Bold} denotes the highest score and \underline{underline} the second-best among non-oracle methods. $^\dagger$Ground-truth subgoals. $^\ddagger$Numbers from the original papers~\cite{dai2026robomme,liu2026t2mem}, as no public checkpoint is available.}
    \label{tab:main}
  \begin{tabular}{@{}l*{16}{c}c@{}}
    \toprule
    & \multicolumn{4}{c}{Counting} & \multicolumn{4}{c}{Permanence} & \multicolumn{4}{c}{Reference} & \multicolumn{4}{c}{Imitation} & \\
    \cmidrule(lr){2-5}\cmidrule(lr){6-9}\cmidrule(lr){10-13}\cmidrule(lr){14-17}
    & Bin & Pick & Swing & Stop & Video & Button & Video & Button & Pick & Video & Video & Video & Move & Insert & Pattern & Route & \\
    Method & Fill & Xtimes & Xtimes & Cube & Umsk & Umsk & UmskS & UmskS & HighL & Repick & PlcBtn & PlcOrd & Cube & Peg & Lock & Stick & Avg. \\
    \midrule
    \multicolumn{18}{@{}l}{\emph{Frozen \groundsg policy}} \\
    Oracle subgoals$^{\dagger}$ & 83.3 & 100.0 & 99.3 & 34.0 & 99.3 & 96.7 & 98.7 & 77.3 & 54.7 & 98.0 & 100.0 & 98.7 & 88.0 & 20.7 & 94.7 & 58.0 & 81.3 \\
    Gemini subgoals~\cite{dai2026robomme}$^\ddagger$ & 26.0 & 18.0 & 4.0 & 3.0 & 36.0 & 14.0 & 13.0 & 0.0 & 9.0 & 17.0 & 12.0 & 7.0 & 17.0 & 0.0 & 7.0 & 2.0 & 11.6 \\
    Qwen-VL subgoals~\cite{dai2026robomme} & 60.0 & \textbf{92.7} & 6.0 & 0.0 & \textbf{88.0} & 22.7 & 28.7 & 16.7 & 24.7 & \underline{33.3} & 48.0 & 36.0 & 67.3 & 2.0 & 8.0 & 6.7 & 33.8 \\
    MemER~\cite{sridhar2025memer} & \underline{62.7} & 77.3 & 59.3 & 2.7 & 84.0 & \textbf{77.3} & \underline{30.7} & 20.0 & \underline{72.0} & 29.3 & 24.7 & 30.7 & 78.7 & \underline{8.0} & 17.3 & 4.7 & 42.5 \\
    \midrule
    \multicolumn{18}{@{}l}{\emph{Policies trained on RoboMME}} \\
    $\pi_{0.5}$~\cite{intelligence2025pi05} & 28.7 & 36.0 & 34.0 & 8.7 & 21.3 & 20.7 & 20.7 & 9.3 & 17.3 & 0.7 & 32.7 & 28.0 & 28.7 & 2.7 & 4.7 & 8.7 & 18.9 \\
    SAM2Act+~\cite{dai2026robomme}$^\ddagger$ & 40.0 & 76.0 & 25.3 & 0.0 & 27.3 & 32.0 & 18.0 & \textbf{26.7} & 17.3 & 5.3 & 24.7 & 20.0 & 29.3 & 0.0 & 0.0 & 0.0 & 21.4 \\
    TokenDrop+Modul~\cite{dai2026robomme} & 31.3 & 86.7 & 88.0 & 5.3 & 27.3 & 24.7 & 27.3 & 20.0 & 25.3 & 24.0 & 49.3 & \underline{42.0} & 62.7 & 6.0 & 33.3 & 53.3 & 37.9 \\
    FrameSamp+Modul~\cite{dai2026robomme} & 43.3 & \underline{91.3} & \textbf{92.0} & \underline{46.0} & 32.7 & 22.0 & 25.3 & 17.3 & 22.0 & 30.0 & \underline{56.0} & \underline{42.0} & \textbf{81.3} & 5.3 & \underline{54.7} & \underline{65.3} & 45.4 \\
    TTT+Expert~\cite{dai2026robomme} & 27.3 & 35.3 & 36.7 & 4.0 & 34.7 & 22.0 & 17.3 & \underline{20.7} & 20.7 & 8.0 & 24.7 & 24.7 & 41.3 & 1.3 & 2.0 & 2.7 & 20.2 \\
    RMT+Modul~\cite{dai2026robomme}$^\ddagger$ & 33.3 & 60.8 & 37.8 & 4.7 & 31.1 & 11.8 & 17.8 & 2.4 & 17.1 & 4.2 & 32.0 & 31.1 & 24.7 & 2.2 & 3.8 & 8.0 & 20.2 \\
    T$^2$Mem~\cite{liu2026t2mem}$^\ddagger$ & 60.7 & \underline{91.3} & \underline{90.7} & \textbf{70.0} & \textbf{88.0} & \underline{59.3} & 29.3 & 20.0 & 50.7 & \textbf{38.7} & 35.3 & 30.0 & \underline{80.7} & \textbf{36.0} & 50.0 & \textbf{78.7} & \underline{56.8} \\
    \midrule
    \method (ours) & \textbf{75.3} & 83.3 & 87.3 & 22.0 & \underline{85.3} & 45.3 & \textbf{52.0} & 15.3 & \textbf{82.0} & 16.0 & \textbf{80.7} & \textbf{82.7} & 74.0 & \underline{8.0} & \textbf{67.3} & 40.7 & \textbf{57.3} \\
    \bottomrule
  \end{tabular}
\end{table*}

\paragraph{Subgoal selection and object grounding.}
At every decision step $t>D$, \method selects the grounded subgoal $g_t=(c_t,u_t)$ with multiple-choice questions of the form of \cref{eq:answer}.
The first question, $q^{\mathrm{sub}}$, asks which subgoal the robot should be executing now, with the candidates as options:
\begin{equation}
  c_t = \operatorname*{arg\,max}_{c^n\in\mathcal{C}}\; p_\theta\!\left(c^n \mid S_t, x_{t-W+1:t}, \ell, q^{\mathrm{sub}}, \mathcal{C}\right).
  \label{eq:subgoal}
\end{equation}
The answer depends on the history: which steps have already been completed, how many times a step has been repeated, and in which order earlier events occurred.

If the selected subgoal is unchanged, $c_t=c_{t-1}$, the grounded subgoal is kept, $g_t=g_{t-1}$, and no second question is asked.
If $c_t$ is a new subgoal that refers to an object, such as ``lift the cup at $\langle y,x\rangle$,'' a second question asks the decision-making VLM to select its location $u_t$. Specifically, a pointing model (Appendix~\ref{app:grounding}) proposes candidate object locations in the current frame,
\begin{equation}
  \mathcal{U}_t = \mathrm{Point}(x_t) = \{u_t^1, u_t^2, \dots\},
  \label{eq:points}
\end{equation}
and each $u_t^k$ is drawn on the frame with a letter, which gives a marked frame $\tilde{x}_t$.
The question $q^{\mathrm{obj}}(c_t)$ asks which marked object the subgoal $c_t$ refers to:
\begin{equation}
  \begin{aligned}
  u_t = \operatorname*{arg\,max}_{u_t^k\in\mathcal{U}_t}\; p_\theta\bigl(u_t^k \mid{} & \tilde{S}_t, \tilde{x}_{t-W+1:t}, \ell,\\[-2pt]
  & q^{\mathrm{obj}}(c_t), \mathcal{U}_t\bigr),
  \end{aligned}
  \label{eq:object}
\end{equation}
where the tilde indicates that the current frame is replaced by $\tilde{x}_t$ on the copy that reads the question, so that $\tilde{S}_t=f_\theta(S_{t-1},\tilde{x}_t)$ and the persistent state is updated with $x_t$.
The pointing model sees a single frame and proposes every object in it; which of them $c_t$ means, for example the container that hides the blue cube, is decided by \method from its memory.

The location $u_t$ is a coordinate in the image of step $t$ and is kept until the selected subgoal changes.
Over an episode with $T$ decision steps, the decision-making VLM therefore processes $D+T$ frames and answers $T$ subgoal questions, plus one object question each time the selected subgoal changes to one that refers to an object; with $c_D=\varnothing$, the first decision step always selects a new subgoal.
\section{Training Data}
\label{sec:data}

The decision-making VLM answers a question by scoring a given set of options (\cref{eq:jev}), so a training example needs a question, its options and the correct option in addition to the instruction and the frames.
Manipulation datasets provide instructions, observations and actions, but not decisions in this form; we propose a procedure that converts robot demonstrations into such multiple-choice decision data and describe it below.

We use the 1{,}600 demonstrations of the RoboMME training split~\cite{dai2026robomme}, which annotates every demonstration with the subgoal that is being executed at each control step.
We generate the questions from these annotations by fixed rules, without manual labeling: for a frame sampled from a demonstration, a sentence template is filled with the options and the correct answer that the annotations give at that frame.
For instance, the template ``Has the robot already finished this step: $c$?'' receives the answer ``yes'' when the annotated interval of subgoal $c$ ends before the sampled frame.
The annotations and simulator records are used only to construct the questions and their answers; the model receives the instruction, the frames up to the sampled one, and the question.
Validation episodes are used only to measure question accuracy, and test episodes only for closed-loop evaluation.
The questions fall into two groups: decision questions, which are the questions asked during execution, and memory questions, which ask about events that are no longer visible.

\paragraph{Decision questions.}
The first group contains the two questions that \method answers at deployment (\cref{sec:method:decisions}), asked along the demonstrations: 29.5K subgoal questions, whose answer is the subgoal annotated for that step, and 4.0K object questions, whose answer is the marked object at the target position recorded by the simulator (Appendix~\ref{app:questions}).

\paragraph{Memory questions.}
The second group asks about events that are no longer visible, so that the state has to carry them.
It contains 12.3K questions about the video demonstration, such as which movement was shown first, on which target the cube was placed the second time, or which cube was picked up, and 1.8K object questions asked after the containers have been swapped.
In addition, questions about the progress of the episode are not stored in advance but sampled anew at every training step from the subgoal annotations: which subgoal is due, whether a given step is finished, how many times a step has been completed, which step was finished most recently, and which step the demonstration showed first.

\paragraph{Question sampling.}
The stored decision and memory questions form a pool of 47.6K questions. Training samples are drawn from this pool and from the generated progress questions; half of the generated questions are chosen so that the event that determines the answer lies at least $W$ frames before the question.
The question templates and the counts by suite are given in the supplementary material (Appendix~\ref{app:questions}).
\section{Experiments}
\label{sec:exp}

\begin{figure*}[t]
  \centering
  \captionsetup{skip=3pt}
  \input{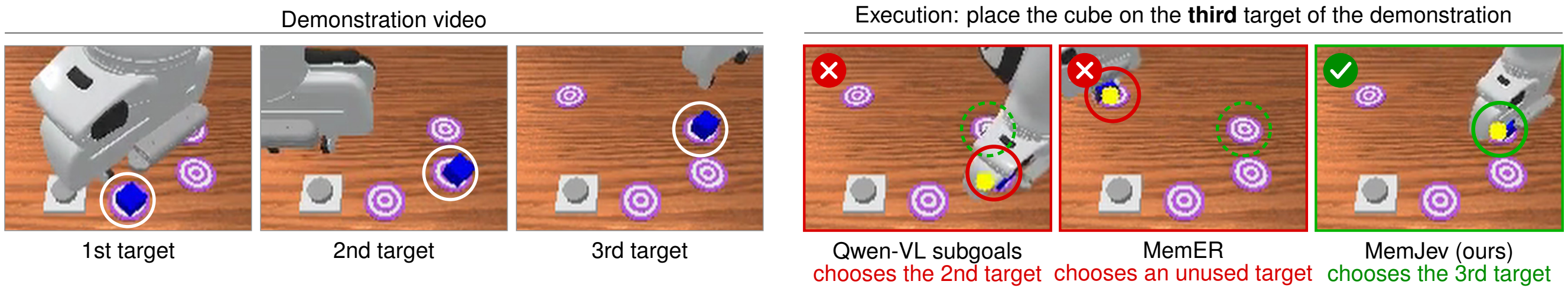}
  \vspace{-10pt}
  \caption{\textbf{Qualitative comparison on a VideoPlaceOrder test episode.}
  The instruction asks for the third target of the demonstration (white rings); each high-level model chooses a target (solid ring; dashed: correct) and the same low-level policy places the cube there, so both failures come from the high-level choice.}
  \vspace{-10pt}
  \label{fig:qual}
\end{figure*}

We evaluate \method on three questions: whether it raises closed-loop success over existing methods (\cref{sec:exp:main}), what a decision costs compared with other high-level models (\cref{sec:exp:latency}), and whether the recurrent state is used as memory (\cref{sec:exp:ablation}).
Comparisons of recurrent update rules and of training duration are given in the supplementary material.

\subsection{Experimental setup}
\label{sec:exp:setup}

\paragraph{Benchmark and metrics.}
RoboMME~\cite{dai2026robomme} contains 16 manipulation tasks organized into Counting, Permanence, Reference, and Imitation suites.
We evaluate 50 test episodes per task, with a maximum of 1{,}300 control steps per episode, and report success rates by suite and averaged across tasks.
Our results in \cref{tab:main} are averaged over three seeds.
Baselines with released checkpoints are evaluated in the same environment as \method. Beyond RoboMME, we also evaluate \method on a real robot; the setup and results are given in Appendix~\ref{app:real}.

\paragraph{Models and execution.}
All evaluations of \method use the released \groundsg checkpoint with fixed weights, a $\pi_{0.5}$~\cite{intelligence2025pi05} policy fine-tuned on RoboMME to follow grounded subgoals~\cite{dai2026robomme}.
The high-level model has 9B parameters and selects subgoals and target objects, with pretrained Qwen3-VL models~\cite{bai2025qwen3vl} supplying candidate points without fine-tuning (Appendix~\ref{app:grounding}).
A frame is taken every $H=16$ control steps, in the demonstration video and during execution, and a decision is made at every frame of the execution using the recurrent state and the last $W=8$ frames.
The high-level modules use camera images, the instruction, and a demonstration video when provided, without access to simulator state during evaluation.

\paragraph{Baselines.}
We distinguish systems that use the frozen \groundsg{} policy from methods that train their own memory-conditioned policies.
In the first group only the source of the grounded subgoal changes, which isolates the high-level decisions: Oracle subgoals passes the ground-truth subgoal and gives the success that the policy reaches with correct decisions, Gemini and Qwen-VL subgoals generate the subgoal with a VLM, and MemER generates it from retained keyframes.
The second group provides context for the broader benchmark.
Rows marked with $^\ddagger$ in \cref{tab:main} are taken from the original papers; all other rows are evaluated by us.
For TokenDrop, TTT and RMT we show the variant with the highest average in~\cite{dai2026robomme}.

\subsection{Comparison with baselines}
\label{sec:exp:main}

\Cref{tab:main} reports 57.3\% mean success for \method over three seeds, against 33.8\% for Qwen-VL subgoals and 42.5\% for MemER with the same low-level policy and test episodes; our method ranks first or second on 8 of the 16 tasks among the non-oracle methods.
The gain over MemER, the stronger of the two baselines, is largest in the Reference, Imitation and Counting suites (26.2, 20.3 and 16.5 points), whose decisions depend on events that lie many frames back, such as the order of the demonstrated placements, a demonstrated pattern or the number of repeated motions.
These are also the suites in which rebuilding the state from the recent frames lowers success most (\cref{sec:exp:ablation}), which indicates that the gain comes from the history carried in the recurrent state.
\Cref{fig:qual} shows such an episode, where the correct target depends on the order of placements in the demonstration: Qwen-VL subgoals and MemER select other targets, and ours selects the correct one.
The policy-training results are reported as context because their architectures, supervision, and adapted parameters differ from ours.

\subsection{Decision latency}
\label{sec:exp:latency}

\begin{table}[t]
  \centering
  \small
  \setlength{\tabcolsep}{2.5pt}
  \caption{\textbf{Decision latency.} Memory: memory computation of one decision (ms), and Late mem.\ the same at the end of an episode; Decision: all model calls of one decision (median, ms); Episode: all model calls of one episode (s); Succ.: test success (\%).}
  \resizebox{\columnwidth}{!}{%
  \begin{tabular}{@{}lccccc@{}}
    \toprule
    Method & Memory & Late mem. & Decision & Episode (s) & Succ. \\
    \midrule
    Qwen-VL subgoals & 1{,}193 & 1{,}757 & 1{,}161 & 25.3 & 33.8 \\
    MemER & 1{,}752 & 2{,}387 & 1{,}797 & 39.4 & 42.5 \\
    \method (ours) & \textbf{113} & \textbf{111} & \textbf{158} & \textbf{19.7} & \textbf{57.3} \\
    \bottomrule
  \end{tabular}}
  \vspace{-10pt}
  \label{tab:latency}
\end{table}

\Cref{tab:latency} times all models of a method on one GPU: the memory computation of one decision (Memory), the same in the last 100 control steps of an episode (Late mem.), and all model calls of one decision (Decision, median) and of one episode.
The memory computation of \method takes 113\,ms per decision and does not grow over an episode, whereas the subgoal calls of Qwen-VL subgoals and MemER grow with the episode.
Our median decision takes 158\,ms, 7.3 and 11.4 times less than those of Qwen-VL subgoals and MemER.
Object questions occur in about 12\% of the decisions and call the pointing models; summed over an episode, all model calls of \method take 19.7\,s on average, the lowest of the three methods.

\subsection{Is the recurrent state used as memory?}
\label{sec:exp:ablation}

\paragraph{Interventions on the state.}
Question accuracy and task success alone do not establish which historical information the model uses.
We therefore change what the recurrent state contains during evaluation of the final model, keeping its weights, the frozen policy, and the 800 test episodes unchanged (\cref{tab:interventions}).
We test three interventions:
\begin{itemize}
  \item \emph{No memory.} At every decision, the recurrent state is cleared and rebuilt from only the most recent eight frames. The model therefore keeps the recent window but no earlier history.
  \item \emph{Memory frozen.} Execution frames are not written into the state, so the state holds only the demonstration. The last seven executed frames remain visible through the attention window.
  \item \emph{Swap.} Before each decision, the state is replaced by the state of a different episode of the same task, taken at the corresponding point.
\end{itemize}

\begin{table}[t]
  \centering
  \footnotesize
  \setlength{\tabcolsep}{4pt}
  \caption{\textbf{Memory interventions on the final model.}
    Test success (\%), averaged over the 16 tasks (Avg.) and per suite; Full is the unchanged state.}
  \begin{tabular}{@{}lccccc@{}}
    \toprule
    Condition & Avg. & Counting & Permanence & Reference & Imitation \\
    \midrule
    Full (Ours) & \textbf{57.3} & \textbf{67.0} & \textbf{49.5} & \textbf{65.3} & \textbf{47.5} \\
    No memory & 26.8 & 16.0 & 39.0 & 26.0 & 26.0 \\
    Memory frozen & 39.5 & 13.0 & 42.0 & 57.0 & 46.0 \\
    Swap & 36.6 & 42.0 & 45.0 & 28.0 & 31.5 \\
    \bottomrule
  \end{tabular}
  \vspace{-10pt}
  \label{tab:interventions}
\end{table}

All three interventions lower mean success, and for each of them the episodes solved only by the full model outnumber those solved only under the intervention (Appendix~\ref{app:interventions}).
Resetting the state at each decision lowers success from 57.3\% to 26.8\%, with lower scores in every suite.
Removing executed frames lowers Counting from 67.0\% to 13.0\%, whereas the other three suites decrease less.
Replacing the state with that of another episode lowers success to 36.6\%.
These measurements show that changing the contents of the state lowers closed-loop success and that the affected suites depend on which frames are removed.

\paragraph{Memory training and the recurrent state.}
\Cref{tab:ab22train} trains the 9B model for 20{,}000 steps with the recurrent memory on or off and with or without the memory questions, keeping the schedule and the number of questions per step fixed; with memory off, every question sees only the last eight frames, in training and at test.
Training without the recurrent memory lowers test success from 57.3\% to 38.2\%, with the largest drop in Reference (65.3\% to 38.0\%).
Removing the memory questions lowers success by 9.7 points with memory and by 4.7 points without it, so these questions help mainly when the state can carry the events they ask about.
The model without memory is tested in the condition it was trained for, so its lower success is not a mismatch between training and test; without the memory questions, Reference falls the most.

\begin{table}[!t]
  \centering
  \small
  \setlength{\tabcolsep}{3.5pt}
  \caption{\textbf{Training with and without memory.}
    Test success (\%) when the model is trained with or without the recurrent memory (Memory) and with or without the memory questions (Mem.\ Q.); without memory, every question sees only the last eight frames.}
  \resizebox{\columnwidth}{!}{%
  \begin{tabular}{@{}ccccccc@{}}
    \toprule
    Memory & Mem.\ Q. & Avg. & Counting & Permanence & Reference & Imitation \\
    \midrule
    \checkmark & \checkmark & \textbf{57.3} & 67.0 & \textbf{49.5} & \textbf{65.3} & \textbf{47.5} \\
    \checkmark & -- & 47.6 & \textbf{68.0} & 41.0 & 42.5 & 39.0 \\
    -- & \checkmark & 38.2 & 41.5 & 38.5 & 38.0 & 35.0 \\
    -- & -- & 33.5 & 39.0 & 35.5 & 30.0 & 29.5 \\
    \bottomrule
  \end{tabular}}
  \vspace{-10pt}
  \label{tab:ab22train}
\end{table}

\begin{figure}[t]
  \centering
  \captionsetup{skip=1pt}
  \includegraphics[width=\columnwidth,trim={0 6pt 0 4pt},clip]{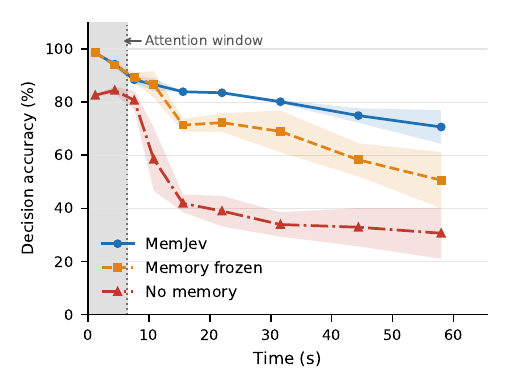}
  \caption{\textbf{Decision accuracy over time.} Once the evidence is outside the attention window, \method still selects the correct subgoal in most decisions, while accuracy falls when the state is not updated during execution (Memory frozen) or is rebuilt from the last eight frames (No memory).}
  \label{fig:distance}
\end{figure}

\paragraph{Retention of past evidence.}
We ask whether the state holds old frames: for the nine tasks whose evidence is shown in a video before the robot acts, \cref{fig:distance} groups the decisions of the test episodes by the time since that video ended and reports how often the selected subgoal matches the simulator's.
While the end of the video is inside the eight-frame window, the first 6.4\,s, \method and Memory frozen select the correct subgoal in at least 94.0\% of the decisions.
Beyond the window, No memory falls to 30.6\% at 51 to 65\,s after the video, whereas \method stays at 70.6\% and Memory frozen at 50.6\%.
Memory frozen matches \method for the first 12\,s and then falls behind, so later decisions also depend on the frames written during execution.

\section{Conclusion}
\label{sec:conclusion}

We introduced \method, in which a decision-making VLM with recurrent visual memory selects grounded subgoals for a frozen low-level policy.
We also proposed a training pipeline that gives this model memory with decision and memory questions built from robot demonstrations.
On RoboMME, \method reaches 57.3\% mean success, 14.8 points above the best method with the same policy, while its memory cost per decision stays constant and its median decision takes 158\,ms.
Memory for history-dependent tasks can thus be learned at the high-level decision interface without changing low-level control.

{
    \small
    \bibliographystyle{ieeenat_fullname}
    \bibliography{main}
}

\clearpage
\setcounter{page}{1}
\twocolumn[\centering\Large Supplementary Material\\\vspace{1.0em}]

\appendix
\crefname{appendix}{Appendix}{Appendices}
\Crefname{appendix}{Appendix}{Appendices}
\crefname{subappendix}{Appendix}{Appendices}
\Crefname{subappendix}{Appendix}{Appendices}
\crefname{subsubappendix}{Appendix}{Appendices}
\Crefname{subsubappendix}{Appendix}{Appendices}

\section{Implementation Details}
\label{app:rationale}
\label{app:implementation}

\subsection{Recurrent update}
\label{app:gdn}
\method keeps the recurrent state of the 24 Gated DeltaNet layers~\cite{yang2025gdn} in the decision-making VLM, and the other 8 of the 32 decoder layers, every fourth layer, use full attention.
A $256{\times}256$ frame occupies 64 visual tokens.
Each Gated DeltaNet layer keeps, for each of its 32 heads, a matrix $\mathbf{M}\in\mathbb{R}^{d_k\times d_v}$ with $d_k=d_v=128$ that stores key--value associations.
For the $i$-th token, with query $\mathbf{q}_i$, key $\mathbf{k}_i$ and value $\mathbf{v}_i$ computed from its hidden state $\mathbf{h}_i$ (queries and keys are $\ell_2$-normalized), the layer computes
\begin{align}
  \bar{\mathbf{M}}_i &= \alpha_i\, \mathbf{M}_{i-1}, \label{eq:gdn-decay}\\
  \mathbf{M}_i &= \bar{\mathbf{M}}_i + \beta_i\, \mathbf{k}_i \bigl(\mathbf{v}_i - \bar{\mathbf{M}}_i^{\top} \mathbf{k}_i\bigr)^{\!\top}, \label{eq:gdn-write}\\
  \mathbf{r}_i &= \mathbf{M}_i^{\top} \mathbf{q}_i \,/ \sqrt{d_k}, \label{eq:gdn-read}
\end{align}
where $\alpha_i = \exp\!\bigl(-e^{a}\,\mathrm{softplus}(\mathbf{w}_a^{\top}\mathbf{h}_i + b)\bigr)\in(0,1)$ is a decay gate with one learned pair $(a,b)$ per head and $\beta_i=\sigma(\mathbf{w}_b^{\top}\mathbf{h}_i)$ is a write strength.
\Cref{eq:gdn-decay} forgets, \cref{eq:gdn-write} moves the value stored under key $\mathbf{k}_i$ toward $\mathbf{v}_i$, and \cref{eq:gdn-read} is the output of the layer for this token.
The state $S_t$ of \cref{eq:state} is the collection of these matrices over all layers and heads after the last token of frame $x_t$, and $f_\theta$ applies \cref{eq:gdn-decay,eq:gdn-write} once per token of the frame.
The state has $24\cdot 32\cdot 128\cdot 128\approx 1.26\times10^{7}$ entries, 48.00\,MiB in 32-bit floats. The attention cache of the instruction and the last $W$ frames occupies about 20.7\,MiB.
The decay parameters $(a,b)$ are trained directly (Appendix~\ref{app:training}).

\subsection{Training details}
\label{app:training}

\paragraph{Trained parameters.}
The rank-16 LoRA adapters~\cite{hu2021lora} ($\alpha=32$, dropout 0.05) are placed on the projection matrices of all layers of the language model: the input, gate and output projections of the recurrent layers, the query, key, value and output projections of the full-attention layers, and the projections of the feed-forward blocks.
The decay parameters $(a,b)$ of the recurrent layers (Appendix~\ref{app:gdn}) are trained directly, and the vision encoder and all other weights are frozen.

\paragraph{Optimization.}
We use AdamW without weight decay, a learning rate of $10^{-4}$ with linear warm-up over the first 100 steps and cosine decay, and gradient clipping at norm 1.
Training runs for 20{,}000 steps of 16 questions.
Examples of similar length are processed together with right padding, which leaves the answer of each example unchanged. A batch holds at most 12{,}288 tokens per GPU.
The model runs in bfloat16, with the recurrence and the decay parameters in 32-bit floats.
Training takes 7.3 hours on four B300 GPUs.

\subsection{Subgoal candidates and object selection}
\label{app:grounding}

\paragraph{Subgoal predictor.}
The subgoal predictor $\rho_\psi$ is a Qwen3-VL model~\cite{bai2025qwen3vl} with rank-16 LoRA, initialized from the grounded-subgoal predictor released with RoboMME and trained for five epochs on the 1{,}600 instruction--plan pairs of the training episodes.
For a step that the instruction does not determine, such as which demonstrated motion comes next, it lists every subgoal that can fill the step, and we discard the order of its output.

\paragraph{Pointing model.}
The object question is defined in \cref{sec:method:inference} (\cref{eq:points,eq:object}).
$\mathrm{Point}$ uses pretrained Qwen3-VL models~\cite{bai2025qwen3vl} without fine-tuning and runs only when the selected subgoal changes to one that refers to an object, which is the case in about 12\% of the decisions and 3.1 times per episode on average on the test split.
$\mathrm{Point}$ upscales the frame from 256 to 512 pixels, proposes a point for every object, drops a proposal that lies within 10 pixels of a point already kept, orders the points from left to right and keeps at most 26.
The selected location $u_t$ is written into the subgoal sentence as $\langle y,x\rangle$ in the coordinates of the $256{\times}256$ frame.

\section{Question Construction}
\label{app:questions}

\Cref{tab:supp:questions} lists the training questions, which are built from episodes of the RoboMME training split. Each question starts with the sentence \emph{``The images are the robot's front-camera views so far, one every 16 control steps, oldest first; the last one is the current view.''}

\begin{table*}[t]
  \centering
  \small
  \setlength{\tabcolsep}{5pt}
  \caption{\textbf{Training questions.} The first two groups form the pool of 47{,}619 questions, and the third group is generated during training. The ordinal in the demonstration questions varies.}
    \label{tab:supp:questions}
  \begin{tabular}{@{}llp{8.3cm}lr@{}}
    \toprule
    Group & Question & Text & Options & Count \\
    \midrule
    Decision & Subgoal & \emph{Which subgoal should the robot be executing now?} & candidates $\mathcal{C}$ & 29{,}487 \\
     & Object & \emph{In the last image the objects are marked with letters. The robot's next subgoal is ``$c$''. Which marked object does this subgoal refer to?} & marks & 4{,}030 \\
    \midrule
    Memory (stored) & Object after a swap & The object question, asked after the containers have been swapped & marks & 1{,}792 \\
     & Demonstrated movement & \emph{In the demonstration video shown at the beginning, what was the robot's first movement?} & movements & 9{,}910 \\
     & Demonstrated object & \emph{In the demonstration video shown at the beginning, which marked cube did the robot pick up?} / \emph{\ldots on which marked target was the cube placed the second time?} & marks & 2{,}400 \\
    \midrule
    Memory & Current subgoal & \emph{Which subgoal should the robot be executing now?} & all subgoals & -- \\
    (generated) & Completion & \emph{Has the robot already finished this step: ``$c$''?} & yes / no & -- \\
     & Count & \emph{How many times has the robot finished the step ``$c$'' so far?} & 0--5 & -- \\
     & Previous step & \emph{Which step did the robot finish most recently?} & all subgoals & -- \\
     & Demonstration order & \emph{In the demonstration video shown at the beginning, which step was shown first?} & all subgoals & -- \\
    \bottomrule
  \end{tabular}
\end{table*}

\paragraph{Labels of decision questions.}
The RoboMME training set labels each step of a demonstration with its subgoal, and we use this label as the ground-truth answer to the subgoal question at that step.
Episodes whose instruction has only one candidate subgoal are not used for subgoal questions, since every such question would have a single option.
For an object question, the pointing model marks the frame, and the answer is the mark within 12 pixels of the target position recorded by the simulator. A question without such a mark is not used.

\paragraph{Labels of memory questions.}
For an object question after a swap, the episode is replayed in the simulator to recover the position of each container, and the answer is the mark within 12 pixels of the container that the subgoal refers to.
For the question about the robot's movement in the demonstration video, the answer is the $k$-th demonstrated subgoal in the RoboMME annotations, with consecutive repeats merged.
For the questions about objects in the demonstration video, the episode is replayed in the simulator to find which cube was picked up or on which target the cube was placed, and the answer is the mark within 12 pixels of that object in the current frame.
The memory questions generated during training are derived from the subgoal annotations of the episode.

\paragraph{Composition.}
By suite, the pool contains 13{,}954 questions from Counting tasks, 8{,}204 from Permanence, 10{,}688 from Reference and 14{,}773 from Imitation.
During training a sample is drawn from the pool with probability 0.7 and is a generated question otherwise, and half of the generated questions are chosen so that the event that determines the answer lies at least $W=8$ frames before the question.

\section{Additional Evaluation}
\label{app:evaluation}

All closed-loop results in this section use the RoboMME test split: 16 tasks with 50 episodes each and at most 1{,}300 control steps per episode.
The low-level policy is the frozen \groundsg checkpoint, and question accuracy is measured on 1{,}566 validation questions.
C, P, R, and I denote the Counting, Permanence, Reference, and Imitation suites, and Avg.\ is the mean success rate over the 16 tasks.
The intervention rows of \cref{tab:interventions}, the ablation rows of \cref{tab:ab22train} and the training-duration results of \cref{fig:length} are single evaluation runs. In this run, the final model reaches 56.9\%, and the 57.3\% of \cref{tab:main,tab:interventions,tab:ab22train} is its mean over three seeds.

\subsection{Memory interventions}
\label{app:interventions}

No memory, Memory frozen and Swap are defined in \cref{sec:exp:ablation}, and a fourth condition, \emph{No demo}, does not write the frames of the demonstration video into the state.
Every condition lowers the mean success of the full model (\cref{tab:interventions}). Under No demo, success is 33.9\%: Counting, whose tasks have no demonstration video, stays at 68.5\%, while Permanence falls to 21.5\%, Reference to 35.0\% and Imitation to 10.5\%.
For every condition, the episodes solved only by the full model outnumber those solved only under the condition (259 against 18 for No memory, 153 against 14 for Memory frozen, 187 against 25 for Swap and 224 against 40 for No demo).
When attention covers only the current frame, the model trained on our memory questions answers 91.1\% of the memory questions with the recurrent state and 33.7\% without it, whereas the public checkpoint answers 35.7\% and 31.2\%, so the state carries the history only after our training.

\subsection{Latency and memory footprint}
\label{app:cost}

\Cref{tab:latency} is measured with one process on one RTX PRO 6000 GPU at batch size 1, with warm-up calls excluded, on two test episodes of each of the 16 tasks.
Writing a frame into the state and answering a question on the copied state each take about 55\,ms, so the memory computation of a decision takes about 110\,ms. A decision that asks an object question also calls the pointing models, and the subgoal predictor runs once per episode.
FrameSamp+Modul, which keeps its memory inside the policy and makes no high-level decision, takes 207\,ms per decision on average and reaches 45.4\% success.

\Cref{tab:supp:cost} times one 4B model on one B300 GPU as the episode grows.
Answering a question from the state and the last 8 frames takes 44\,ms at 128 frames, against 303\,ms when every frame is read again for the question, and the state carried between questions stays at 73\,MB, which is the recurrent state plus the cache of the eight frames in the window.
Keeping every frame's KV instead answers in the same 44\,ms but grows to 362\,MB at 128 frames and takes 7.3\,s instead of 5.1\,s to read them.

\begin{table}[!t]
  \centering
  \footnotesize
  \setlength{\tabcolsep}{3pt}
  \caption{\textbf{Cost of keeping the state.} One 4B model on one B300 GPU, batch 1, warm-up excluded. Re-read: the frames are processed again for each question. Time per question: answering one question once the episode's frames have been read. Kept (MB) and Peak (GB): the state carried between questions and the peak GPU memory, at 128 frames. Val acc.: validation accuracy (\%) of the same condition.}
    \label{tab:supp:cost}
  \begin{tabular}{@{}lccccccc@{}}
    \toprule
    & \multicolumn{4}{c}{Time per question (ms)} & Kept & Peak & Val \\
    \cmidrule(lr){2-5}
    Frames in the episode & 16 & 32 & 64 & 128 & (MB) & (GB) & acc. \\
    \midrule
    Last 8 frames, re-read & 40 & 40 & 39 & 39 & 0 & 9.4 & 75.9 \\
    All frames, re-read & 61 & 92 & 157 & 303 & 0 & 11.0 & 90.5 \\
    All frames, KV kept & 39 & 39 & 41 & 44 & 362 & 10.1 & 90.5 \\
    \method{} (state + last 8) & 42 & 39 & 41 & 44 & 73 & 9.4 & 89.6 \\
    \bottomrule
  \end{tabular}
\end{table}

\subsection{Training ablation of memory and memory questions}
\label{app:ab22train}

This section gives the details of \cref{tab:ab22train}.
With memory off, both training and test use only the last eight frames for each question, so a model trained without memory is tested in the condition it was trained for. Without the memory questions, training uses only the subgoal and object questions asked during execution, with the same number of questions per step.
Removing the recurrent memory lowers success by 19.1 points when the memory questions are used (57.3\% to 38.2\%) and by 14.1 points when they are not (47.6\% to 33.5\%).
Accuracy on the 1{,}566 validation questions follows the same order as test success (90.3\%, 87.5\%, 83.9\% and 82.7\% for the four rows of the table).
Without the memory questions, Reference falls the most (65.3\% to 42.5\%).
The first row is the final model.

\subsection{Effect of the recurrent update}
\label{app:memory}

The recurrent layers of the backbone use the Gated DeltaNet (GDN) update~\cite{yang2025gdn}.
To examine the recurrent update separately from the rest of the system, we replace this update $f_\theta$ in \cref{eq:state} with Gated DeltaNet-2 (GDN-2)~\cite{hatamizadeh2026gdn2} or Kimi Delta Attention (KDA)~\cite{kimi2025linear}, while keeping the training questions, learning-rate schedule, and 5{,}000-step training budget fixed.
We measure accuracy on 1{,}566 validation questions and task success on the test split.
This comparison tests whether a better fit to recorded questions also produces more successful control.
At initialization, the GDN-2 and KDA models reproduce the answers of the GDN model at initialization, with a KL divergence of 0.0002 between their answer distributions.
Summed over the 24 recurrent layers, the persistent recurrent state occupies 48.00\,MiB for GDN, GDN-2, and KDA (49.50\,MiB including the short-convolution state).
The full-attention cache for the eight-frame window occupies about 20.7\,MiB and is the same for all update rules.
All three update rules use the same LoRA adapters (43.3M parameters). With the parameters of the recurrent layers that are trained directly, the trained parameters number 43.3M for GDN, 68.4M for KDA and 118.8M for GDN-2.

\begin{table}[t]
  \centering
  \small
  \setlength{\tabcolsep}{3pt}
  \caption{\textbf{Recurrent updates after 5{,}000 training steps.}
    Q.\ acc.\ is accuracy (\%) on 1{,}566 validation questions, averaged over the 16 tasks.
    Success is the test success rate (\%), and C, P, R, and I are the corresponding suite success rates.}
    \label{tab:memory}
  \begin{tabular}{@{}lcccccc@{}}
    \toprule
    Update rule & Q.\ acc. & Success & C & P & R & I \\
    \midrule
    GDN~\cite{yang2025gdn} & 86.1 & 46.4 & 61.5 & 38.0 & 49.0 & 37.0 \\
    GDN-2~\cite{hatamizadeh2026gdn2} & 76.9 & 33.8 & 37.5 & 33.0 & 31.5 & 33.0 \\
    KDA~\cite{kimi2025linear} & 83.9 & 44.1 & 61.0 & 37.5 & 40.0 & 38.0 \\
    \bottomrule
  \end{tabular}
\end{table}

For the three update rules, question accuracy and test success give the same order (\cref{tab:memory}): GDN reaches 86.1\% accuracy and 46.4\% success, KDA 83.9\% and 44.1\%, and GDN-2 76.9\% and 33.8\%.
The difference between GDN and KDA lies in Reference (49.0\% \vs 40.0\%), while their other suite scores differ by at most 1.0 point.

\subsection{Effect of training duration}
\label{app:length}

\begin{figure}[t]
  \centering
  \resizebox{\columnwidth}{!}{\input{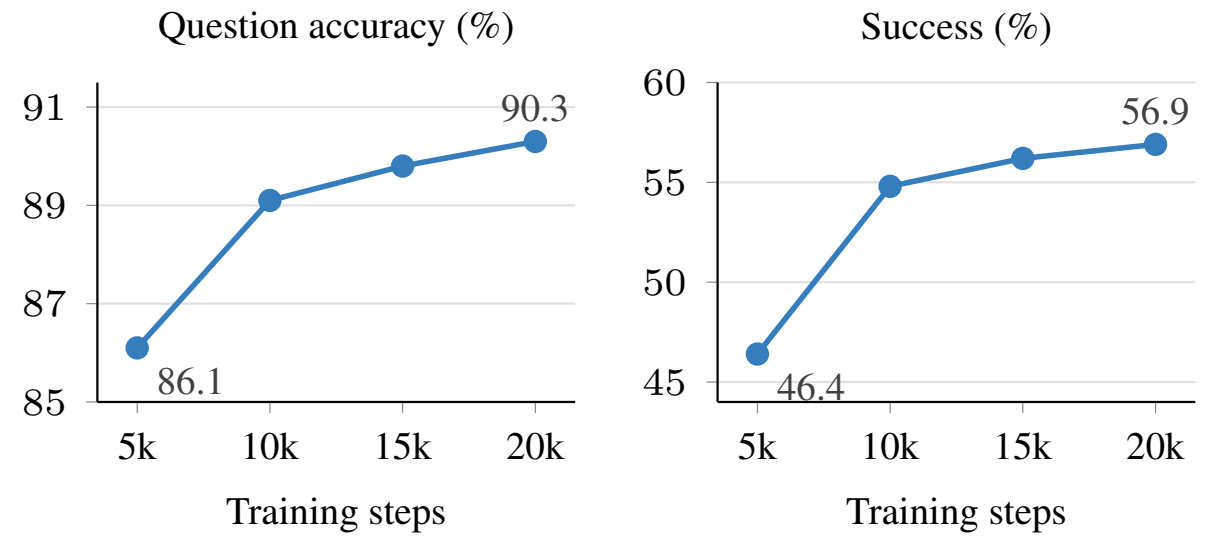}}
  \caption{\textbf{Effect of training duration.} Validation question accuracy and test success of checkpoints from one training run.}
  \label{fig:length}
  \vspace{-20pt}
\end{figure}

We next evaluate GDN checkpoints from the same training run at increasing numbers of optimization steps (\cref{fig:length}).
Validation question accuracy rises from 86.1\% at 5{,}000 steps to 89.1\% at 10{,}000, 89.8\% at 15{,}000 and 90.3\% at 20{,}000 steps.
Test success rises from 46.4\% at 5{,}000 steps to 54.8\% at 10{,}000 and 56.2\% at 15{,}000, and changes little afterwards, reaching 56.9\% at 20{,}000 steps.
Most of the gain between 5{,}000 and 10{,}000 steps comes from Reference (49.0\% to 66.0\%) and Imitation (37.0\% to 45.5\%).

\subsection{Model size and question wording}
\label{app:size}

\Cref{tab:supp:size} evaluates two further conditions on the same 800 test episodes and the same three seeds as \cref{tab:main}.

The 4B row trains the public 4B checkpoint of the decision-making VLM~\cite{vev2026} with the same training questions, the same 20{,}000-step schedule, and the same rank-16 LoRA placement as the 9B model.
Its backbone also has 24 recurrent layers with 32 heads and $d_k=d_v=128$, so the persistent state occupies 48.00\,MiB, the size reported for the 9B model in Appendix~\ref{app:gdn}.
Mean success is 56.1\%, so the recurrent state carries the history of an episode at both model sizes, with the same memory footprint.

The wording row changes the questions at evaluation only and leaves the weights of the final 9B model unchanged.
Each question is rewritten in one of three wordings that do not occur in training: the sentence that describes the frames and the question sentence are both replaced, the option strings are kept, and the wording of an episode is chosen by a hash of its task and index.
One of the three asks \emph{``What subgoal is the robot supposed to be working on at this moment?''} in place of \emph{``Which subgoal should the robot be executing now?''}.
Accuracy on the 1{,}566 validation questions is 90.2\% under the rewritten wordings, against 90.3\% with the training wording, and mean success is 56.8\%.

\begin{table}[h]
  \centering
  \small
  \setlength{\tabcolsep}{4pt}
  \caption{\textbf{Model size and question wording.} Test success (\%) as the mean over three seeds, with the suite means in the last four columns.}
    \label{tab:supp:size}
  \begin{tabular}{@{}lccccc@{}}
    \toprule
    Condition & Success & C & P & R & I \\
    \midrule
    \method, 9B (final model) & 57.3 & 67.0 & 49.5 & 65.3 & 47.5 \\
    \method, 4B & 56.1 & 63.0 & 51.3 & 66.5 & 43.5 \\
    9B, unseen question wording & 56.8 & 67.5 & 49.5 & 64.5 & 45.8 \\
    \bottomrule
  \end{tabular}
\end{table}

\subsection{Qualitative results}
\label{app:qual}

\begin{figure*}[t]
  \centering
  \captionsetup{skip=3pt}
  \input{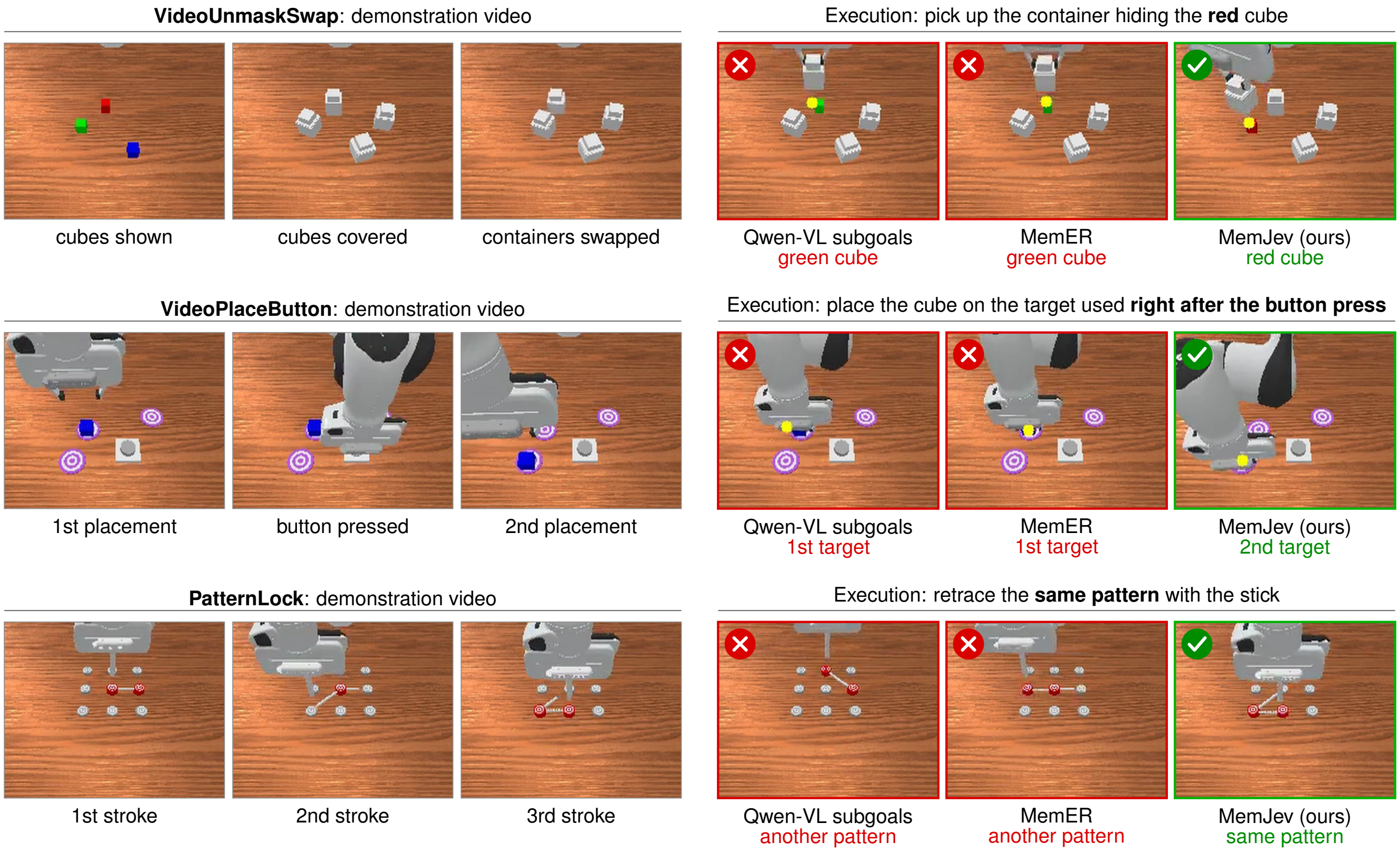}
  \caption{\textbf{Further RoboMME test episodes.}
  Each row gives frames of the demonstration video (left) and the last frame of the execution with each high-level model (right), which pass their subgoals to the same low-level policy.}
  \label{fig:supp:qual}
\end{figure*}

\Cref{fig:supp:qual} shows three further test episodes in the format of \cref{fig:qual}, one from each of the Permanence, Reference and Imitation suites.
In each of them the correct decision is determined by the demonstration video: the container that covers the red cube after the swap, the target used right after the button press, and the traced pattern.
\method selects the container, the target and the pattern of the demonstration.
Qwen-VL subgoals and MemER lift the container of the green cube, place the cube on the target of the first placement, and trace other patterns.
Videos of further RoboMME test episodes are given on the project page.

\section{Real-Robot Experiments}
\label{app:real}

We evaluate \method on a real robot with three tasks in which a person first performs a demonstration and the robot then receives an instruction that only the demonstration resolves.
The three tasks test the memory abilities of three RoboMME suites on the robot: following a hidden object through swaps (Permanence), recalling the order of events (Reference), and counting repeated events (Counting).
The high-level model watches the demonstration and selects the cup to pick up, and, as in \cref{sec:method:inference}, the selected subgoal is passed to a frozen low-level policy.

\subsection{Setup}
\label{app:real:setup}

\begin{figure}[t]
  \centering
  \includegraphics[width=.92\columnwidth]{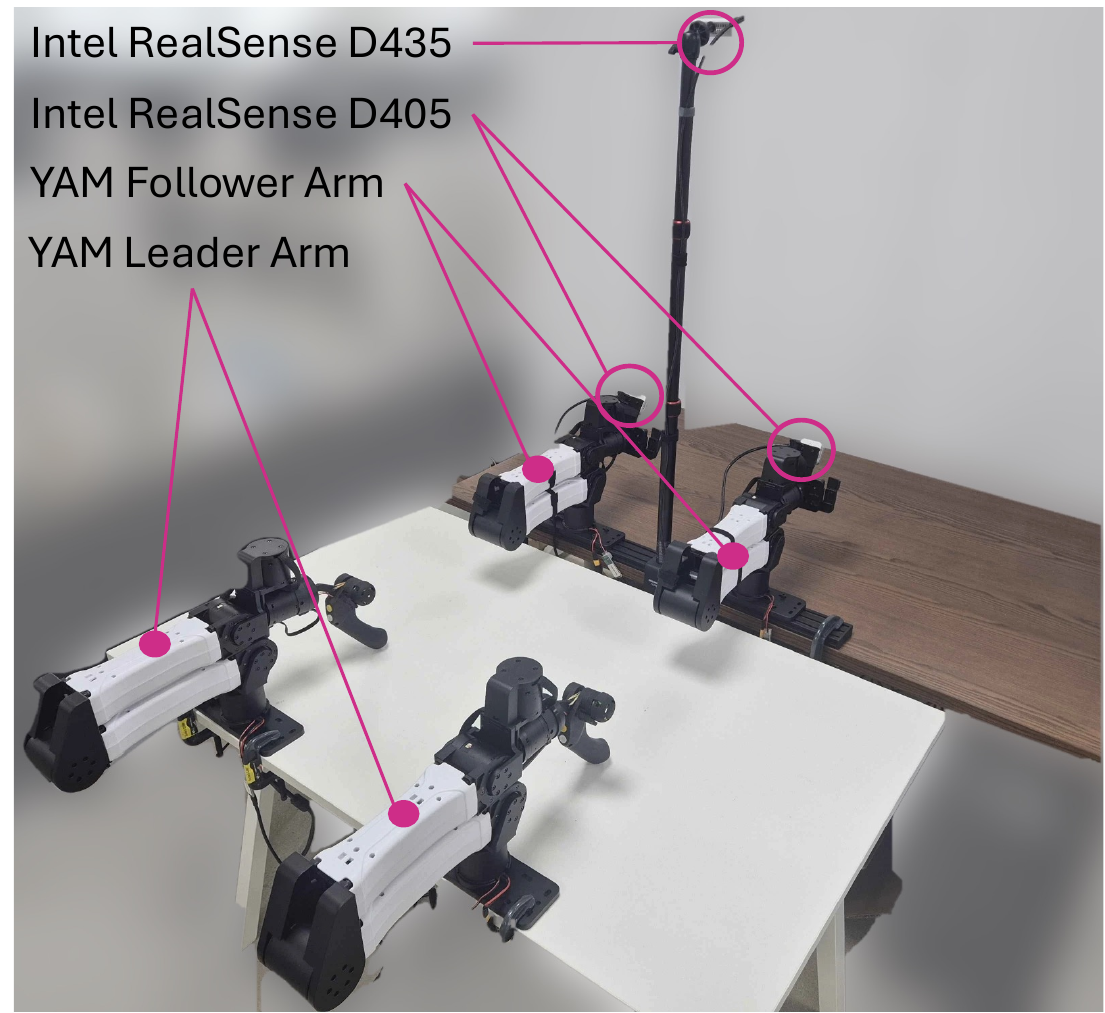}
  \caption{\textbf{Real-robot setup.}
  Two YAM follower arms with wrist cameras, two leader arms for teleoperation, and the top camera whose frames the high-level model receives.}
  \vspace{-20pt}
  \label{fig:real:setup}
\end{figure}

\paragraph{Robot and low-level policy.}
\Cref{fig:real:setup} shows the setup: two YAM follower arms with parallel grippers, each with an Intel RealSense D405 camera on the wrist, two leader arms for teleoperation, and an Intel RealSense D435 camera above the table, which records $640{\times}480$ frames at 30\,Hz.
The right follower arm executes the tasks, and the high-level model receives the frames of the top camera.
Three identical cups stand in a row in front of the robot. Left, middle and right refer to the camera image.
The low-level policy is a $\pi_{0.5}$ model~\cite{intelligence2025pi05} fine-tuned on teleoperated episodes to pick up the cup named in the subgoal.
It receives only the current observation and the subgoal, without any history, and its weights are fixed in all experiments. The same policy executes the decisions of all high-level models, so only the source of the subgoal differs between the compared methods.

\paragraph{Tasks.}
\begin{itemize}
  \item \emph{Shell game.} Two coffee capsules, navy and bronze, are covered by two of the three cups, and the person swaps the positions of the cups several times. The instruction asks for the cup that hides the capsule of a given color.
  \item \emph{Pick-and-drop.} The person lifts the cups one at a time and puts each one back, three to seven lifts in total, and a cup can be lifted more than once. The instruction asks for the cup that was lifted $k$-th.
  \item \emph{Tap count.} The person taps the cups with a spoon, three to nine taps in total. The instruction asks for the cup that was tapped a given number of times.
\end{itemize}
When the instruction is given the three cups look the same, so the correct cup is determined by the demonstration, most of which lies outside the attention window.

\begin{figure*}[t]
  \centering
  \captionsetup{skip=3pt}
  \input{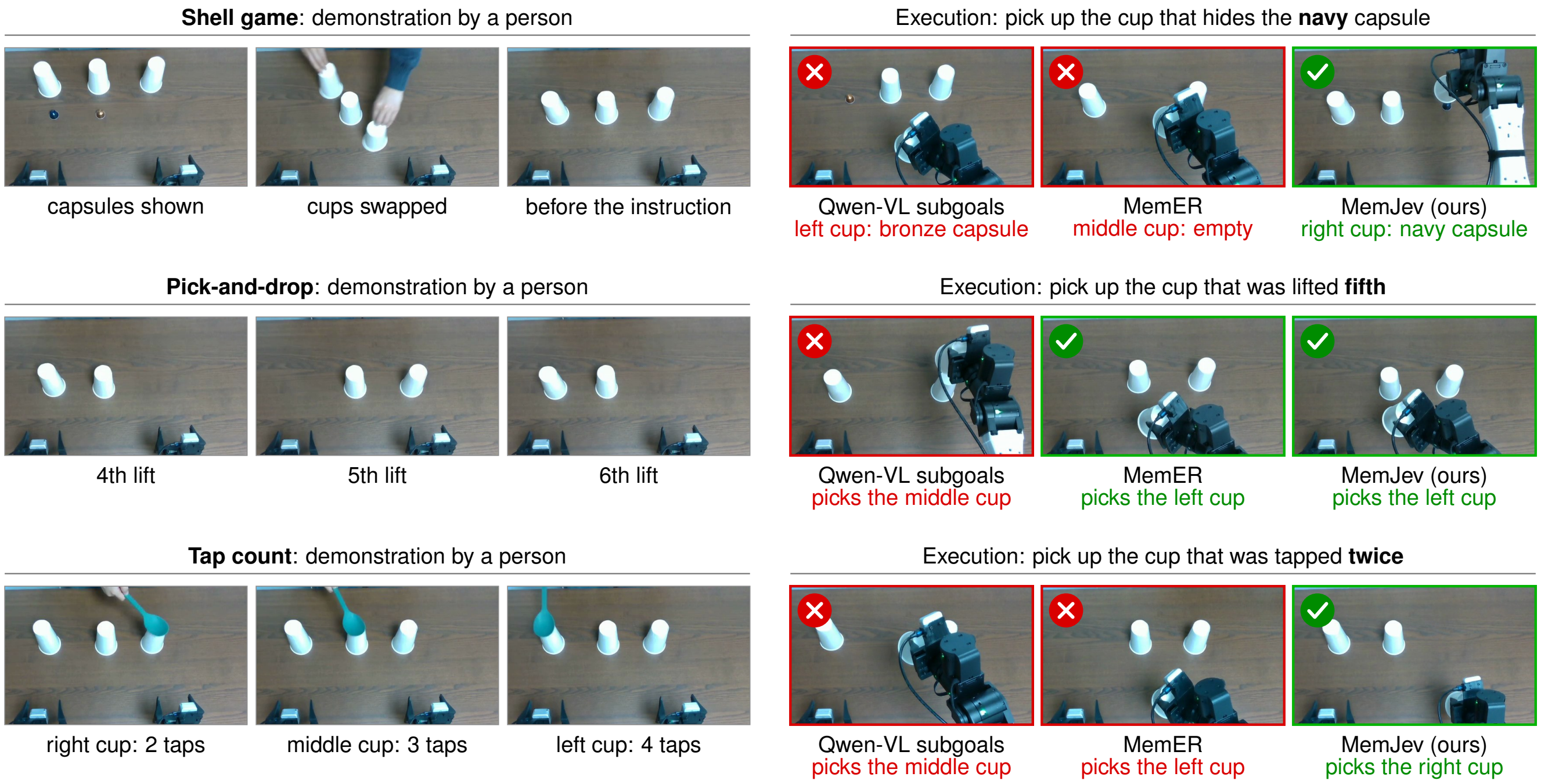}
  \caption{\textbf{Real-robot experiments.}
  In each row a person gives the demonstration (left), and each high-level model then selects the cup that the same low-level policy picks up (right).}
  \label{fig:real:qual}
\end{figure*}

\paragraph{Demonstrations and questions.}
We recorded 50 demonstrations of each task with the top camera and labeled them by fixed rules on the images: the lifts from the pixels at each cup's resting place, the taps from the position of the spoon, and the cup that hides each capsule from the capsule colors before covering and the cup positions after each swap. As in \cref{sec:data}, sentence templates convert the labels into decision questions and memory questions.
The decision question asks which subgoal the robot should be executing now, with the three pick subgoals as options.
The memory questions ask which cup hides a capsule, which cup first covered it, which cup the person lifted $k$-th, and how many times a cup was tapped, at the end of a demonstration and at earlier frames. For pick-and-drop, the answer to the lift-order question selects the subgoal. Each demonstration is used with six temporal offsets of the 0.2\,s sampling.

\paragraph{Models.}
The high-level model has 4B parameters.
It starts from a 4B model trained on RoboMME with the procedure of Appendix~\ref{app:training} and is fine-tuned with LoRA for 4{,}000 further steps on the questions above, with a window of $W=8$ frames, which covers 1.6\,s.
Qwen-VL subgoals and MemER are Qwen3-VL models~\cite{bai2025qwen3vl} trained with rank-16 LoRA in the input formats of their RoboMME implementations~\cite{dai2026robomme}: Qwen-VL subgoals receives ten frames sampled uniformly from the past and the current frame, and MemER receives its retained keyframes and the current window of eight frames.
Qwen-VL subgoals is trained on the same questions as \method, and MemER on the demonstrations of the three tasks in its window format. All three models are trained on the same number of samples.
On the robot the high-level model runs on the RTX 4090 GPU of the robot's computer, next to the low-level policy. With the frames written into the state as they arrive, \method answers 0.13\,s after the last frame of a demonstration and uses 9.6\,GB of GPU memory.

\subsection{Question accuracy on unseen demonstrations}
\label{app:real:heldout}

We first measure whether the selected cup is correct on demonstrations that a model has not seen.
We divide the 50 demonstrations of each task into five groups of ten, and evaluate a model on each group after training it on the remaining 40 demonstrations.

\Cref{tab:real:memory} gives the accuracy of \method over the five groups, with the recurrent state and with the state rebuilt from the last eight frames, and the no-memory condition of \cref{sec:exp:ablation}.
With the state, \method selects the correct cup in 91.5\% of the pick-and-drop questions and 98.5\% of the tap-count questions, and names the cup that first covered a capsule in all 300 questions.
In the shell game, the cup also has to be followed through the swaps, and \method selects the correct cup in 65.0\% of the questions.
Without the state, the accuracy of every question stays at or below 34.8\%, close to the chance level, so the answers are read from the recurrent state.

\begin{table}[!t]
  \centering
  \small
  \setlength{\tabcolsep}{3pt}
  \caption{\textbf{Question accuracy (\%) on unseen real demonstrations.}
    Sum over five folds. No memory rebuilds the state from the last eight frames.}
    \label{tab:real:memory}
  \resizebox{\columnwidth}{!}{%
  \begin{tabular}{@{}lccc@{}}
    \toprule
    Question & Questions & \method & No memory \\
    \midrule
    Shell game: cup to pick up & 100 & \textbf{65.0} & 28.0 \\
    Shell game: first covering cup & 300 & \textbf{100.0} & 32.3 \\
    Pick-and-drop: cup lifted $k$-th & 247 & \textbf{91.5} & 34.8 \\
    Tap count: cup to pick up & 135 & \textbf{98.5} & 31.9 \\
    Tap count: taps on a cup & 150 & \textbf{97.3} & 14.0 \\
    \bottomrule
  \end{tabular}}
\end{table}

\subsection{Comparison with baselines}
\label{app:real:robot}

We compare the three high-level models on 30 shell-game, 50 pick-and-drop and 30 tap-count decisions, all made in trials on the real-robot.
In each trial, a person performs one of the three tasks in front of the robot's top camera, and the video is the demonstration. In the shell game the person hides the two capsules under cups and swaps the cups, in pick-and-drop the person lifts the cups one at a time, and in tap count the person taps the cups with a spoon. None of the three models saw these demonstrations in training.
After the demonstration, the instruction asks the robot to pick up one cup, namely the cup that hides the capsule of a given color, the cup that was lifted $k$-th, or the cup that was tapped a given number of times. The high-level model selects the cup from the demonstration, and the low-level policy picks it up.
A decision is correct when the model selects the cup that the instruction asks for. In the shell game, this is checked by the capsule under the cup that the robot lifts.

\Cref{tab:real:compare} gives the share of correct decisions.
\method selects the correct cup in 80.0\% of the shell-game decisions, 94.0\% of the pick-and-drop decisions and all tap-count decisions.
Qwen-VL subgoals reaches 36.7\%, 84.0\% and 60.0\%, and MemER 33.3\%, 62.0\% and 76.7\%.
MemER issued no pick subgoal in one trial on the robot, which is counted as incorrect, and in one further trial its correct subgoal did not reach the policy, which is counted as correct.
\Cref{fig:real:qual} shows one trial of each task, and \method selects the correct cup in all three.
In the shell game, the instruction asks for the cup that hides the navy capsule after the cups have been swapped. Qwen-VL subgoals selects the cup that hides the bronze capsule and MemER selects an empty cup, whereas \method selects the right cup, which hides the navy capsule.
In pick-and-drop, the instruction asks for the cup that was lifted fifth of seven lifts. MemER and \method select this cup, while Qwen-VL subgoals selects the middle cup, which the person lifted last.
In tap count, the instruction asks for the cup that was tapped twice, whose taps are the first and third of nine. Qwen-VL subgoals selects the cup tapped three times, and MemER selects the cup tapped four times, which received the last taps of the demonstration, whereas \method selects the right cup.
Videos of the trials on the robot are given on the project page.

\begin{table}[h]
  \centering
  \small
  \setlength{\tabcolsep}{4pt}
  \caption{\textbf{Comparison on the real-robot tasks.}
    Correct decisions (\%) on 30 shell-game, 50 pick-and-drop and 30 tap-count decisions.}
    \label{tab:real:compare}
  \begin{tabular}{@{}lccc@{}}
    \toprule
    Method & Shell game & Pick-and-drop & Tap count \\
    \midrule
    Qwen-VL subgoals & 36.7 & 84.0 & 60.0 \\
    MemER & 33.3 & 62.0 & 76.7 \\
    \method (ours) & \textbf{80.0} & \textbf{94.0} & \textbf{100.0} \\
    \bottomrule
  \end{tabular}
\end{table}

\end{document}